\documentclass[journal]{IEEEtran}

\usepackage[left=0.625in,right=0.625in,top=0.75in,bottom=1in]{geometry}
\usepackage{ifpdf}
\usepackage{cite}
\usepackage{array}
\usepackage{dblfloatfix}
\usepackage{url}
\usepackage{amsmath}
\usepackage{ragged2e}
\usepackage{graphicx}
\usepackage{verbatim}
\usepackage{float}
\usepackage{lipsum}
\usepackage{multicol, blindtext}
\usepackage{caption}
\usepackage{subcaption}
\usepackage{xcolor}
\definecolor{mypink2}{RGB}{0, 0, 255}
\definecolor{green}{RGB}{0, 128, 0}
\usepackage{multirow}
\usepackage{array}
\usepackage{tabularx}
\usepackage[acronym]{glossaries}
\usepackage[ruled,vlined]{algorithm2e}
\usepackage{algorithmic}

\usepackage[utf8]{inputenc}
\usepackage{tcolorbox}
\usepackage{xcolor}

\usepackage{url}
\usepackage{hyperref}

\begin{document}

\title{\fontsize{12pt}{12pt}\selectfont Advancing
Autonomous Emergency Response Systems: A Generative AI Perspective}

\author{ Yousef~Emami,~\IEEEmembership{Senior Member,~IEEE,}
        Radha~Reddy,
        Azadeh~Pourkabirian,~\IEEEmembership{Member,~IEEE,}
        and~Miguel~Gutierrez Gaitan,~\IEEEmembership{Senior Member,~IEEE,}
       
\thanks{Copyright (c) 2025 IEEE. Personal use of this material is permitted. However, permission to use this material for any other purposes must be obtained from the IEEE by sending a request to pubs-permissions@ieee.org.}

\vspace{-20pt}}

\maketitle
\begin{abstract}
Autonomous Vehicles (AVs) are poised to revolutionize emergency services by enabling faster, safer, and more efficient responses. This transformation is driven by advances in Artificial Intelligence (AI), particularly Reinforcement Learning (RL), which allows AVs to navigate complex environments and make critical decisions in real time. However, conventional RL paradigms often suffer from poor sample efficiency and lack adaptability in dynamic emergency scenarios. This paper reviews next-generation AV optimization strategies to address these limitations. We analyze the shift from conventional RL to Diffusion Model (DM)-augmented RL, which enhances policy robustness through synthetic data generation, albeit with increased computational cost. Additionally, we explore the emerging paradigm of Large Language Model (LLM)-assisted In-Context Learning (ICL), which offers a lightweight and interpretable alternative by enabling rapid, on-the-fly adaptation without retraining. By reviewing the state of the art in AV intelligence, DM-augmented RL, and LLM-assisted ICL, this paper provides a critical framework for understanding the next generation of autonomous emergency response systems from a Generative AI perspective.
\end{abstract}


\IEEEpeerreviewmaketitle
\section{Introduction}

Autonomous vehicles (AVs) are poised to transform emergency services by enabling faster, safer, and more intelligent responses. Uncrewed Aerial Vehicles (UAVs), as key enablers within the AV ecosystem, provide rapid deployment and precise mobility. They can serve as both aerial base stations and data collectors, enhancing connectivity and information gathering for AV operations. With advanced sensing technologies, real-time data processing, and autonomous decision-making, AVs can function without human intervention, making them a promising solution for disaster relief and emergency response. These self-driving vehicles can navigate hazardous conditions, transport essential supplies and equipment, and evacuate individuals from affected areas with minimal risk to human operators. By leveraging advanced technologies such as Artificial Intelligence (AI) and Machine Learning (ML), AVs can interpret sensory data to support rapid decision-making, route optimization, and threat detection. Integrated with AV systems, AI enhances real-time navigation, accelerates emergency responses, and reduces human error through continuous and objective perception \cite{kurunathan2023machine},\cite{li2019board}.

\begin{figure}[!htbp]
    \centering
    \includegraphics[width=0.48\textwidth]{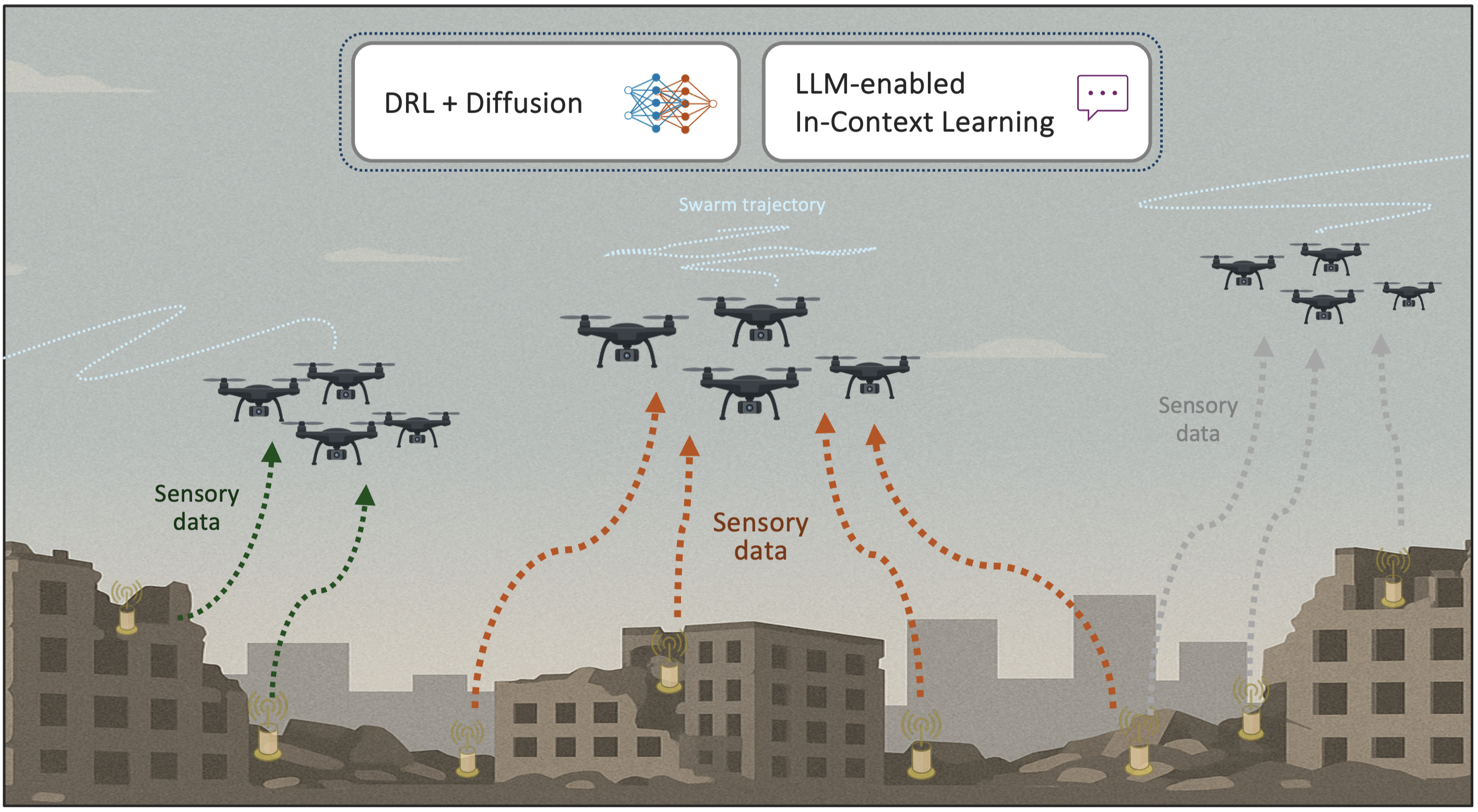}
    \caption{Illustration of public safety UAV scenario, where the UAV follows a trajectory and collects sensory data. LLM-assisted ICL and DM-augmented RL can be used to optimize its operations.}
    \label{drl1080}
\end{figure}

Fig.~\ref{drl1080} shows a public safety scenario where multiple UAVs coordinate their trajectories to collect sensory data from disaster-stricken areas. In this scenario, the UAVs can use Generative AI (GenAI) to optimize data collection by adapting to changing contexts, and enabling intelligent, real-time decision-making directly at the edge \cite{10749978}. This paper discusses two GenAI approaches: Diffusion Model-augmented Reinforcement Learning (DM-augmented RL) and Large Language Model (LLM)-assisted in-context learning (LLM-assisted ICL). 

Although these models are distinct, they provide complementary advancements that together address key limitations in AV decision-making. In the earlier approach, RL provides an adaptive framework that optimizes long-term objectives through continuous interaction with the environment, enabling AVs to navigate complex state and action spaces. However, RL often faces challenges such as the simulation-to-reality gap and high data requirements \cite{li2015energy}. Integrating DMs into RL helps address these issues by generating realistic and diverse synthetic data, thereby improving sample efficiency, policy robustness, and environmental fidelity. Serving as multifaceted enhancers, DMs enable RL to safely simulate rare or high-risk scenarios, capture multi-modal action representations, and support coherent long-horizon planning for complex emergency operations. \cite{emami2025diffusion}.

In the latter approach, LLMs improve the capabilities of AVs by extending their functions beyond autonomous control to include contextual understanding and reasoning aligned with human needs. Through ICL, LLMs can interpret natural language descriptions, real-time sensor data, and operational cues, allowing them to adapt their behavior dynamically without requiring retraining or fine-tuning. This ability helps to bridge the gap between perception and reasoning, enabling LLMs to offer interpretable, language-based guidance to AV systems. Compared to conventional optimization or ML techniques, LLM-assisted ICL simplifies task formulation, improves adaptability, and promotes transparency and user trust.

In this research direction, prior work \cite{emami2025diffusion} investigated the integration of DM with RL and digital twins to improve sample efficiency and mitigate data management challenges. Further, the authors in \cite{emami2025llm} proposed ICL for data collection scheduling as an alternative to Deep Reinforcement Learning (DRL) in emergency scenarios and evaluated its robustness against jailbreaking attacks. Additionally, a framework integrating LLM-enabled ICL with public safety UAVs to address key functions such as path planning and velocity control in emergency response contexts, was introduced in \cite{emami2025prompts}
Together, these DM-augmented RL and LLM-assisted ICL create a foundation for AV systems that are not only technically resilient and efficient, but also context-aware, explainable, and responsive to complex emergencies \cite{cui2024receive}.

In this paper, we first explore how DMs enhance RL by generating synthetic data, refining policy networks, and simulating high-fidelity environments, thereby improving sample efficiency, stability, and robustness. We then analyze the role of LLM-assisted ICL as a lightweight, training-free optimizer that uses natural language commands and contextual feedback to facilitate real-time decision-making for AVs. Finally, we present a comparative analysis of these two paradigms, underscoring their complementary strengths and the challenges that lie ahead. Thus, our \textit{contribution} lies in advancing the next generation of autonomous emergency response systems by investigating DM-augmented RL and LLM-assisted ICL as separate and complementary approaches. We highlight the distinct strengths of each method and their suitability for AVs, providing a clear framework that guides researchers and practitioners.

The remainder of this paper is organized as follows: Section \ref{sec2} provides an overview of AVs. Section \ref{sec3} presents DM-augmented RL, including its merits, and a case study. Section \ref{sec4} examines LLM-assisted ICL, discussing its key merits along with a case study. Section \ref{sec5} offers a comparative discussion of the two paradigms, and Section \ref{sec6} concludes the paper with future directions.

\section{Autonomous Vehicles Overview}
\label{sec2}

AVs represent a transformative advancement with the primary objective of improving road safety, along with the secondary benefits of enhanced fuel efficiency, reduced congestion, and increased mobility. Their operational framework is built on a hierarchical three-tier architecture for perception, planning, and control.  
\par
The foundational perception layer leverages a heterogeneous suite of sensors, including radar, LiDAR, and cameras, to build a comprehensive understanding of the driving environment. The sensory data gathered are subsequently processed within a software architecture using ML and Deep Learning (DL) pipelines, systematically structured into interdependent modules that encompass perception, prediction, planning, and control. 
\par
The planning layer functions as the vehicle's "brain," responsible for high-level decision-making and trajectory planning. This layer is formally modeled as a Markov Decision Process (MDP), for which learning-based approaches, such as RL, are used to solve. Within this MDP framework, RL enables autonomous operation by empowering AVs to optimize trajectories, avoid obstacles, and make adaptive decisions through continuous interaction with the environment. By learning from these interactions to improve performance and achieve long-term objectives, RL serves as the core technology for intelligent decision-making in AVs.  

The last layer, which is the control layer, executes driving decisions by generating physical commands for steering and braking. This system is enhanced by connectivity via Vehicular Ad-hoc Networks (VANETs), which enable V2X communication. V2X enables vehicles to exchange critical data with other vehicles (V2V), infrastructure (V2I), and pedestrians (V2P). A network of Electronic Control Units (ECUs) manages these components, forming the distributed computational backbone that digitally controls everything from braking and steering to entertainment \cite{you2019advanced},\cite{li2022deep}.

The integration of AVs into emergency services relies not on a single technology, but on a powerful synergy of advanced systems. LiDAR and sensors supply high-fidelity and real-time environmental data, which AI and ML algorithms interpret to enable split-second decision-making and precise navigation. Despite this, significant challenges remain, including handling rare edge cases and addressing the ethical complexities of programming for life-or-death decisions. Beyond onboard intelligence, V2X communication allows AVs to coordinate with infrastructure and other vehicles, optimizing response routes. However, the efficiency of this connectivity is directly tied to the availability of advanced infrastructure, which can exacerbate inequities between different regions \cite{li2022internet}.

The formidable challenges in integrating AVs into emergency services, particularly technological limitations in unpredictable environments and the pressing need for public trust, underscore a fundamental shortcoming in existing AI paradigms: Conventional RL models are lacking the robustness and adaptability necessary for life-or-death decision-making. To overcome these limitations, advanced frameworks such as DM-augmented RL have emerged as promising solutions. By generating and training on a vast spectrum of synthetic scenarios, from complex multi-vehicle collisions to hazardous weather conditions, DMs significantly enhance the robustness of policies. This capability enables AVs to maintain reliable performance even in previously unseen situations, effectively addressing technological constraints and establishing the reliability essential for public confidence.

However, DM-augmented RL alone is inadequate to manage the dynamic and unpredictable nature of real-world emergency response. The challenges of system integration and the need for interpretable, adaptive decision-making demands a framework capable of real-time adjustment without extensive retraining. This is precisely where LLM-assisted ICL offers unique advantages. Leveraging their extensive world knowledge and reasoning capabilities, LLMs can serve as high-level cognitive co-pilots, processing contextual information from sensory data to generate reasoned and interpretable guidance. By enhancing transparency, contextual awareness, and adaptability, LLM-assisted ICL not only addresses key integration challenges but also fosters explainable decision-making, an essential step toward building public trust and realizing the full potential of autonomous emergency response systems.

\begin{table*}[!htbp]
\centering
\caption{DM-Augmented RL: Advantages and Relevance to Emergency Autonomous Vehicles}
\label{tab:dm_rl_avs}
\begin{tabular}{|p{4.8cm}|p{5.5cm}|p{6.5cm}|}
\hline
\textbf{Challenge} & \textbf{Advantages of DM-Augmented RL} & \textbf{Relevance to Emergency AVs} \\
\hline
 Complex multi-modal decisions & Captures multi-modal actions; models complex behaviors & Models diverse action distributions, enabling AVs to choose among multiple plausible maneuvers \\
\hline
 Rare safety-critical events & Generates high-quality actions; improves decisions & Augments data with realistic synthetic scenarios, improving preparedness for rare events \\
\hline
 Adherence to safety constraints & Models diverse behaviors; supports conditional generation & Allows guided sampling with safety conditions for safer decision-making \\
\hline
 Generalization to new tasks & Enables iterative trajectory refinement; long-horizon control & Multitask conditioning enables AVs to generalize to novel situations and missions \\
\hline
Learning from human experts & State-of-the-art offline RL performance; enhances behavior cloning & Supports better imitation learning via expressive policy representation \\
\hline
Multi-agent interaction & Stable training; resilient to varying conditions & Models uncertain behaviors of other agents, improving coordination and robustness in swarm or traffic scenarios \\
\hline
\end{tabular}
\end{table*}

\section{Diffusion Model-Augmented Reinforcement Learning} 
\label{sec3}

The deployment of RL in AVs faces several salient technical challenges. A major barrier is low sample efficiency; RL agents cannot learn through millions of trial-and-error interactions in real-world transportation scenarios. When learning from pre-collected operational data (offline RL), standard algorithms are often ineffective because their simplistic policies cannot capture the complex, multi-modal decision-making of  AVs. Moreover, these systems suffer from data scarcity, fail to cover the vast number of potential driving situations, and are insufficient for training robust policies in high-dimensional and unpredictable transportation systems.

DMs provide a foundational upgrade to RL in the domain of AVs, offering key advancements across three dimensions:
\begin{itemize}
    \item \textbf{Data Augmentation:} DMs act as powerful data synthesizers, learning the distribution of historical driving data to generate vast, high-fidelity synthetic experiences. This mitigates data scarcity and low sample efficiency, enabling the development of robust policies without the need for costly data collection in the real world.

    \item \textbf{Generative Planning:} As non-autoregressive planners, DMs can produce entire multi-step trajectories in a single inference. This approach reduces compounding errors and enables more temporally coherent long-horizon planning, which is critical for navigating complex environments.

    \item  \textbf{World Modeling:} DMs' ability to model complex, high-dimensional distributions facilitates the creation of realistic generative simulations. These risk-free environments provide a high-fidelity training ground that narrows the simulation-to-reality gap, a vital step for validating safety-critical systems.
\end{itemize}
        
For emergency AVs operating in unpredictable, high-stakes environments, DMs are vital for robust RL. Table \ref{tab:dm_rl_avs} summarizes how DM-augmented RL addresses key AV challenges where conventional RL falls short. First, by modeling diverse action distributions, DM-augmented RL enables the complex, multimodal decision-making required to adapt to dynamic crises. Second, DMs overcome the scarcity of real-world data on rare, critical events by generating high-fidelity synthetic edge cases for comprehensive training. Furthermore, DMs ensure safety compliance through guided sampling, enforcing operational and ethical constraints during policy generation. Finally, their capacity for multitasking allows AVs to generalize learned competencies across diverse emergencies, minimizing the need for retraining. 

Additionally, learning from human experts, a cornerstone of trustworthy AV design, is strengthened through the expressive policy representations provided by DMs, which enable more precise imitation and behavior aligned with humans. In multi-agent settings where AVs must coordinate with other vehicles, responders, or pedestrians, DM-augmented models further excel by capturing the uncertain and probabilistic nature of others’ behaviors. Collectively, these capabilities position DM-augmented RL as a promising paradigm for developing autonomous emergency vehicles that are intelligent, adaptive, and capable of operating safely and reliably under the most challenging real-world conditions.

Beyond emergency response, DM-augmented RL proves particularly effective for complex network optimization and high-dimensional control tasks. One of its defining advantages is enhanced expressiveness; DMs can represent complex, multi-modal action distributions far more effectively than traditional Gaussian policies, enabling AVs to capture diverse behavioral modes and navigate intricate solution spaces. This expressiveness also yields higher-quality samples, as DMs iteratively refine action generation to improve decision accuracy and system performance in resource allocation and control optimization. Moreover, DM-augmented RL demonstrates flexibility and adaptability, as it can model a wide range of behaviors and incorporate conditional information such as network states or constraints. It also provides inherent planning capabilities, allowing for iterative trajectory refinement and effective long-horizon decision-making. In offline RL scenarios, DM-based approaches achieve state-of-the-art performance on benchmarks such as D4RL, improving both behavior cloning and policy optimization. They also deliver consistent and stable convergence across varying initial conditions and random seeds, surpassing conventional DRL methods in terms of robustness and reliability.

\subsection{Case Study on UAV Swarm Coordination Task}

\begin{figure*}[!htbp]
    \centering
    \includegraphics[width=0.9\textwidth]{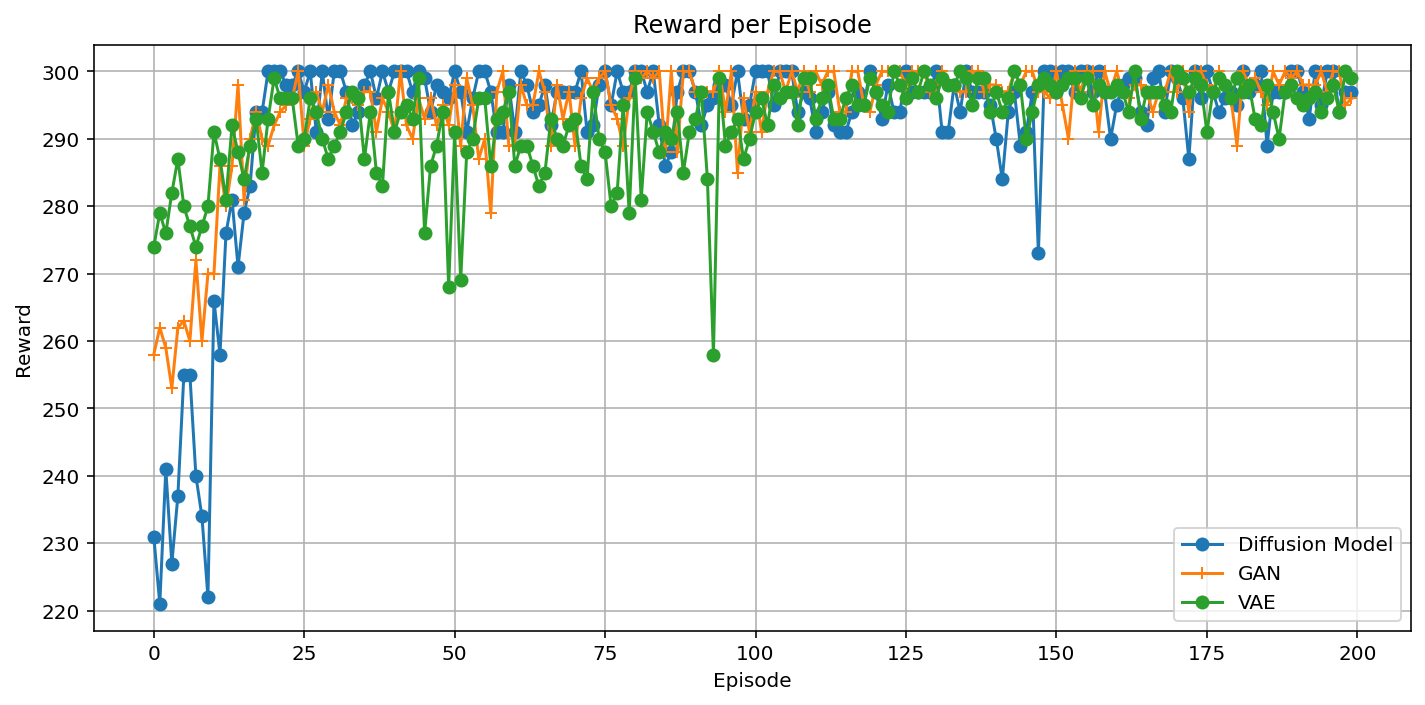}
    \caption{Comparison of generative model performance in coordinating a 4-UAV swarm: DMs, GANs and VAEs.}
    \label{drl1060}
\end{figure*}

In this case study, DMs were evaluated against two prominent generative approaches, Variational Autoencoders (VAEs) and Generative Adversarial Networks (GANs). In a swarm coordination task involving four UAVs, DMs outperformed both GANs and VAEs. As illustrated in Fig.~\ref{drl1060}, DMs achieved a higher peak reward (300), converged more rapidly (~ within 100 episodes), and exhibited greater training stability, with a reward variance of $\sigma = 2.1$, compared to GANs (280, $\sigma = 8.7$) and VAEs (260, $\sigma = 5.3$). The iterative denoising mechanism of DMs enabled them to capture complex multi-agent dynamics more effectively for velocity prediction. In contrast, GANs were hampered by training instability, while VAEs were limited by the over-regularization of their latent space.

By effectively modeling complex multi-agent dynamics, DMs generate realistic velocity predictions that balance inter-agent spacing and swarm cohesion, making them the most reliable choice for safety-critical, real-world applications of UAVs. GANs and VAEs, although helpful, are hindered by training instability and limitations in the latent space, highlighting the critical role of model selection in multi-agent AI systems.

Overall, DM-augmented RL embodies a trade-off between computational cost and policy quality. Although this approach requires greater computational resources and careful hyperparameter tuning, its ability to generate expressive, high-quality actions, solve complex optimization problems, and maintain stability in offline RL makes it particularly valuable in scenarios where solution quality and adaptability are crucial. Despite slower inference, the advantages of enhanced policy expressiveness, robust performance, and superior final outcomes often justify the added complexity and computational overhead \cite{du2024enhancing, emami2025diffusion, zhu2023diffusion, 10812969}. Nevertheless, even with DM-augmented RL, dynamic, uncertain, and context-rich environments, such as those encountered in real-world emergency response require not only high-quality actions but also real-time adaptability, flexibility, and interpretable reasoning. This is where LLM-assisted ICL becomes indispensable.

\section{LLM-Assisted In-Context Learning} 
\label{sec4}

The integration of LLMs into autonomous systems represents a paradigm shift, moving beyond traditional modular designs toward end-to-end, learning-driven architectures. By leveraging techniques like multimodal tokenization and transformer-based attention, these systems achieve a unified understanding of their environment, seamlessly fusing diverse data streams (camera, LiDAR, maps) into a cohesive model of the world. This enables unprecedented capabilities in contextual reasoning, flexible decision-making, and intuitive interaction. The emergence of LLMs as general-purpose reasoning engines is powered by the ICL. This paradigm allows a pre-trained model to perform novel tasks by processing a carefully constructed prompt, which includes a natural language instruction and a set of demonstration examples. It operates by furnishing the model with contextual cues within the input prompt, typically a task specification and illustrative examples, that implicitly define the target input-output mapping, thereby bypassing the need for resource-intensive retraining. 
\par
Formally, given a new input query, the LLM evaluates a set of candidate answers and selects the one with the highest likelihood based on the provided context. The effectiveness of ICL depends more on the structural presentation and formatting of these examples than on the absolute correctness of the labels, suggesting that the model is learning the underlying task structure from the demonstrations. This makes ICL a uniquely flexible and efficient method for task adaptation, particularly well-suited for AVs where conditions change rapidly and predefined solutions are insufficient.

The deployment paradigm is crucial for integrating this technology into AVs. An edge server hosts the LLM to provide low-latency access. The operational cycle starts when the AV collects real-time sensory data, and transmits to an edge-based LLM, which converts it into a structured, natural language task description. This description forms the foundation of ICL, explicitly defining the task objective, relevant input data (e.g., sensor readings), operational rules and constraints, expected output format, and feedback mechanism.
\par
Upon processing this detailed, context-rich prompt, the LLM infers an appropriate action or plan, which the AV subsequently executes. The resulting performance, captured through metrics such as packet loss reduction or variations in energy consumption, is then measured and fed back into the system. Through this iterative feedback loop, the system learns from experience, dynamically refining the LLM's decisions to form a closed, loop, adaptive control system—all without any retraining \cite{emami2025llm}.
\par
The optimization framework provides a clear and efficient mechanism for real-time adaptation by integrating objective, externally computed performance metrics with LLM-driven contextual reasoning. Unlike conventional RL, where learning depends on parameter updates, this system performs iterative optimization within the context space, allowing the LLM to refine decisions dynamically through prompt-based feedback rather than retraining. Each task leverages observable, ground-truth-independent cost metrics derived directly from AV or edge measurements, ensuring operational transparency and deployability in real-world settings. By clearly separating numerical evaluation from reasoning, this framework creates a well-posed, interpretable, and training-free feedback loop. This ensures the responsiveness, explainability, and reliability required for autonomous AV-edge collaboration in mission-critical operations.


The LLM-assisted ICL framework optimizes several key functions of AVs. It generates and evaluates discretized trajectories based on operational requirements using feedback from executed routes to iteratively refine navigation. This enables dynamic, cost-efficient discretized trajectory planning. In disaster-response scenarios, speed and adaptability outweigh fine-grained control. The discretized trajectory framework enables real-time, low-latency decision-making, making it well-suited for LLM-assisted reasoning and edge deployment. By reducing computational complexity while preserving strategic flexibility it allows AVs to adapt swiftly to dynamic environments. The approach delegates fine-grained control to onboard systems, forming an efficient interface between high-level cognitive planning and low-level execution. This abstraction is vital for enhancing mission responsiveness, reliability, and overall operational effectiveness.
\par
Meanwhile, for velocity control, the framework dynamically adjusts the AV’s speed by evaluating candidate velocities against real-time traffic conditions. Moreover, the data collection schedule dynamically prioritizes sensors based on their requirements, generating adaptive schedules that minimize data loss and ensure reliable transmission through continuous feedback and iterative improvement. Additionally, the framework manages power control for AVs, optimizing transmit power to ensure reliable data transfer. Overall, LLM-assisted ICL offers a lightweight, training-free, and rapidly adaptable solution, which is crucial for time-sensitive emergency scenarios \cite{emami2025prompts}.

Recent work introduces a novel paradigm for public safety UAVs by using LLMs with ICL as a training-free, adaptive alternative to complex DRL methods. The Flight Resource Allocation scheme based on LLM-Enabled In-Context Learning (FRSICL) \cite{emami2025frsicl} framework demonstrates how an edge-deployed LLM can optimize a UAV’s data collection schedule and flight velocity in real-time, minimizing Age of Information (AoI) for wildfire monitoring through natural language task descriptions and environmental feedback. This is extended in  ICL-based Data Collection Scheduling (ICLDC) \cite{emami2025llm}, which focuses on data collection schedule and addresses security vulnerabilities, including black-box jailbreaking attacks, with a defense based on prompt perplexity analysis.

\subsection{Case Study on Data Collection Schedule and Velocity Control}

\begin{figure} 
    \centering
    \includegraphics[width=8cm, height=6cm]{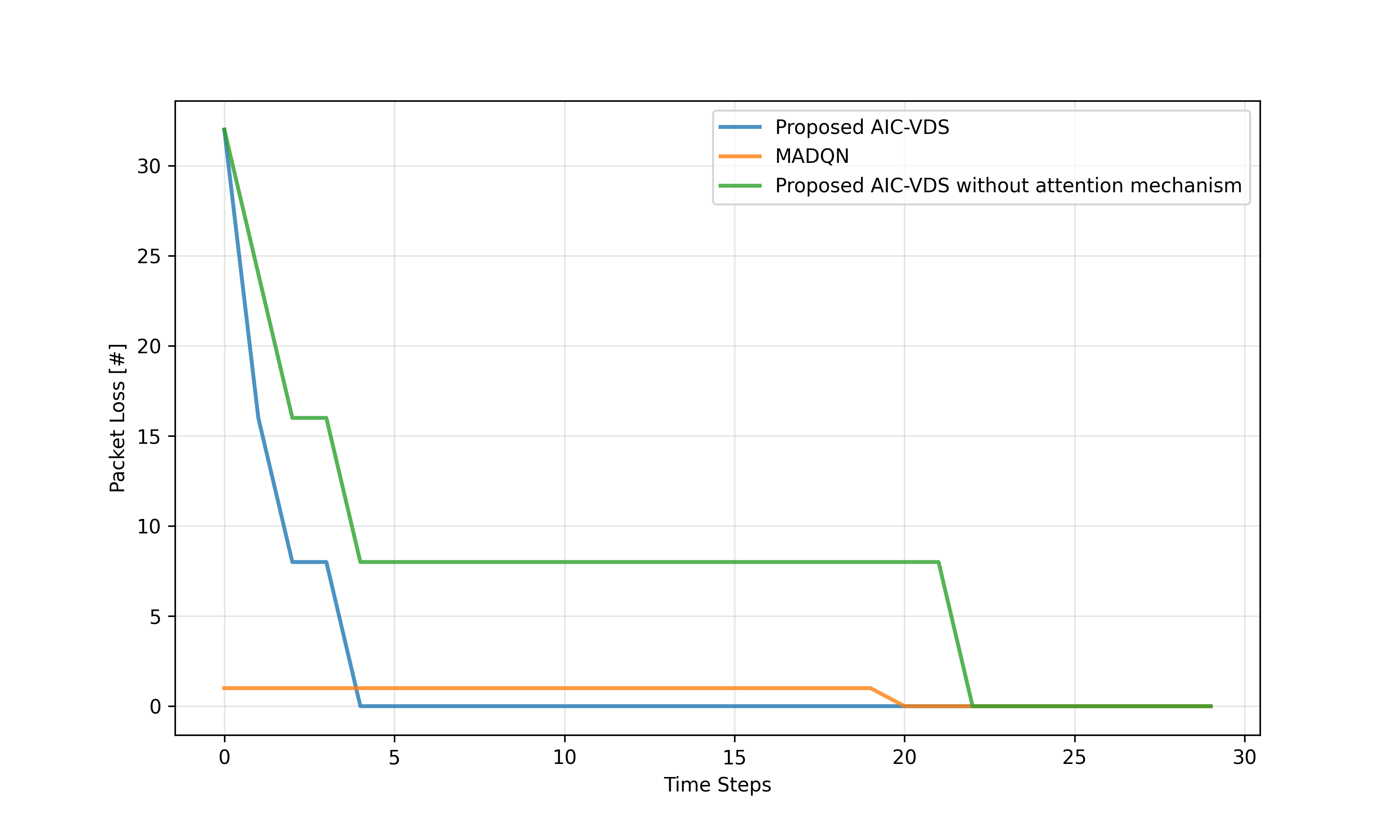}
    \caption{Convergence of the AIC-VDS framework and baselines.}
    \label{drl1020}
\end{figure}

Each UAV autonomously optimizes its data collection schedule and velocity through an attention-based ICL mechanism that dynamically adapts to environmental states, including location, sensor queue length, residual energy, and channel conditions. A core component of this approach is the attention layer, which compresses and prioritizes multi-dimensional input data, allowing the model to focus on the most critical information for decision-making. The primary objective of this process is to minimize packet loss. Fig.~\ref{drl1020} shows the convergence of the proposed AIC-VDS with \texttt{GPT-4O-mini}, compared to the case without the attention mechanism and the Multi-Agent DQN (MADQN) baseline (the performance of MADQN is shown for the last episode). The proposed approach and the MADQN baseline achieve comparable performance, while the case without attention module results in higher packet loss. The proposed approach also achieves faster convergence than the no-attention case.

\begin{table*}[t]
\centering
\caption{Comparison of Optimization Approaches for Autonomous Vehicles}
\label{tab:comparison}
\begin{tabular}{|l|p{6cm}|p{6cm}|}
\hline
\textbf{Feature} & \textbf{DM + RL Hybrid} & \textbf{LLM Optimizer (ICL)} \\
\hline

\textbf{Training Requirements} & 
Generates synthetic data to improve sample efficiency and accelerate pre-training; still computationally heavy due to denoising processes. &
No retraining or parameter updates required; adapts instantly through contextual prompting. \\
\hline

\textbf{Deployment Speed} &
Faster than standalone DRL through pre-trained synthetic environments; inference still limited by model complexity. &
Enables rapid adaptation and deployment in UAV missions through real-time prompt modification. \\
\hline

\textbf{Simulation-to-Reality Gap} &
Generates realistic environments to bridge the sim-to-real gap and enhance robustness. &
Mitigates modeling dependency through natural language task formulation and contextual reasoning. \\
\hline

\textbf{Data Dependency} &
Alleviates data scarcity via synthetic data generation and improved offline RL performance. &
Relies on linguistic descriptions and few-shot demonstrations rather than structured datasets. \\
\hline

\textbf{Adaptability} &
Improved adaptability via conditional generation and multi-modal behavior modeling. &
Highly flexible; adapts to unseen tasks through context without modifying model parameters. \\
\hline

\textbf{Planning and Decision Quality} &
Supports long-horizon planning and refined action generation through iterative denoising. &
Improves decision interpretability and transparency through human-readable reasoning. \\
\hline

\textbf{Computational Complexity} &
Higher than DRL due to multi-network diffusion pipelines and sensitive hyperparameters. &
Low computational cost during inference; main challenge lies in latency for edge/onboard deployment. \\
\hline

\textbf{Key Advantage} &
Bridges sim-to-real gap and enhances policy robustness in complex UAV conditions. &
Lightweight, training-free, and human-interpretable optimization adaptable to new tasks. \\
\hline

\textbf{Key Limitation} &
Computationally intensive; sensitive to noise and parameter tuning; slower inference. &
Convergence and reliability under real-world uncertainty remain unverified. \\
\hline

\end{tabular}
\end{table*}

\section{Comparative Insights} 
\label{sec5}
DM-augmented RL offers a powerful approach to enhance the robustness of autonomous decision-making. As illustrated in Fig. \ref{drl1020}, DMs demonstrate superior performance in multi-UAV coordination tasks, achieving higher stability and convergence compared to GANs and VAEs. This improvement stems from the model’s ability to capture complex, multi-modal action distributions and refine them through iterative denoising. DM-augmented RL offers several key advantages for emergency AVs. Its high expressiveness enables the modeling of intricate AV behaviors, while enhanced action quality allows for the generation of precise, high-value decisions in emergencies. The flexibility of DMs supports the generation of conditions and adaptive behavior modeling, making them suitable for heterogeneous and unpredictable environmental conditions. Moreover, DMs facilitate long-horizon planning by iteratively refining AVs trajectories and exhibit robustness through stable training across varying environmental conditions. They also deliver state-of-the-art performance in offline RL, improving behavior cloning and enabling safer pre-deployment policy learning.
\par
However, these strengths come with notable limitations. The added computation from DMs can hinder scalability in large DRL tasks, such as optimizing citywide communication networks. DM-augmented architectures are computationally intensive due to their multi-step denoising processes, leading to slower inference and higher training costs. Performance is susceptible to hyperparameters, and fine-tuning the number of denoising steps or noise schedules is often critical for stability. Additionally, DM-augmented RL systems can struggle under severe distribution shifts, exhibit slower initial convergence, and involve complex multi-network implementations, which can hinder real-time deployment in constrained UAV platforms. A key limitation of DMs is their sensitivity to data quality. In practical DRL applications—particularly those using real-world network traffic data, the presence of strong noise and outliers can degrade a DM's ability to learn an accurate representation of the data distribution, hindering overall task performance.
\par
Despite these challenges, the combination of DMs with RL presents a compelling path toward bridging the simulation-to-reality gap and enhancing adaptability for safety-critical applications. By generating high-quality synthetic experiences and modeling diverse behavioral distributions, DMs provide DRL frameworks with more realistic, data-efficient learning pipelines. For emergency AVs, this hybrid paradigm enhances decision reliability, situational awareness, and mission resilience —key capabilities for autonomous systems operating in uncertain, time-sensitive environments. Future work should prioritize optimizing inference speed, reducing architectural complexity, and developing adaptive denoising mechanisms to enable rapid, resource-efficient deployment in real-world emergency operations.
\par
Recent advances in prompting methodologies have significantly enhanced LLM capabilities across various applications. By leveraging their fundamental capacity for natural language understanding and generation, LLMs enable a novel approach to complex problem-solving. Traditional computational methods typically demand rigorous problem formalization and algorithmic implementation through specialized systems. In contrast, LLM-assisted approaches permit problem articulation in natural language, allowing the model to progressively generate responses through iterative refinement based on contextual descriptions and prior outputs.
\par
LLMs have evolved from domain-specific language processors into versatile reasoning and optimization engines that can generalize across diverse tasks without additional training. A salient mechanism enabling this capability is ICL, wherein an LLM adapts to new tasks by conditioning on a contextual prompt containing task instructions and demonstration examples. Rather than relying on gradient updates or retraining, the model leverages the semantic structure of the prompt to infer task-relevant relationships, allowing for rapid and adaptive decision-making.
\par
ICL offers a lightweight and efficient alternative to conventional DRL and fine-tuning-based adaptation frameworks by eliminating the need for retraining, reducing computational overhead, and enabling rapid, real-time responsiveness to environmental changes. Its training-free adaptability, low operational complexity, and compatibility with compact edge-deployable models make it especially well-suited for dynamic, resource-constrained applications such as emergency AVs. By leveraging natural-language task descriptions and example-based reasoning instead of complex optimization pipelines, ICL provides a flexible, interpretable, and energy-efficient mechanism for continuous adaptation, thereby bridging the gap between high-level reasoning and on-the-fly decision-making in intelligent autonomous systems. ICL can be extended using attention-based ICL, where an attention module filters and prioritizes input data before sending it to an LLM, allowing the LLM to make AV control decisions without training or fine-tuning. In Intelligent Transportation Systems (ITS), each AV collects large amounts of input data. Instead of feeding all raw data to the LLM which is inefficient and exceeds input limits the attention mechanism assigns importance scores to each sensor and selects only the most critical ones. These compressed, high-priority data are then formatted as natural-language context and passed to the LLM, which performs ICL. This enables rapid adaptation to dynamic environments without reinforcement learning or retraining. In summary, attention reduces input complexity, and the LLM uses the remaining context to reason and generate decisions directly through prompting. The development of Multi-Modal ICL (MM-ICL) represents a critical advancement for AV systems. This approach would empower the underlying Foundation Models (FMs) to process and learn from complex, nested demonstrations that fuse visual (camera, LiDAR) and textual (traffic rules, driver intent, navigation instructions) data within a single prompt.
\par
This LLM-assisted methodology provides several distinct advantages for emergency AVs. The framework enables rapid adaptation to diverse operational scenarios through straightforward modifications to the prompt content, allowing mission parameters to be redefined in real-time without structural model changes. The iterative refinement process enhances output quality through context-aware feedback, enabling continuous improvement without retraining or fine-tuning. Moreover, prompt engineering serves as a low-cost mechanism for model control: by incorporating explicit operational directives, such as safety thresholds, bandwidth limits, or energy constraints, ICL enables AVs to adhere to mission-specific requirements while maintaining flexibility.
\par
Once demonstrations are provided, the model can process new scenarios and infer optimal actions by referencing contextual knowledge embedded within prior examples. In effect, the LLM generalizes decision-making strategies from a limited number of demonstrations, adapting dynamically to unseen conditions without requiring modifications to its internal parameters. This ability to perform few-shot adaptation through language-defined context makes LLMs particularly attractive for emergency AV operations, where pre-training for every possible contingency is infeasible.
\par
The effectiveness of ICL-assisted optimization can be evaluated using operational performance metrics relevant to AVs, such as packet loss rate, data throughput, and AoI. By assessing the degree to which the model’s inferred actions align with optimal outcomes derived from simulation or expert baselines, researchers can measure the accuracy, responsiveness, and adaptability of the LLM’s decision-making. Experimental evidence from recent studies on LLM-assisted optimizers shows that properly designed prompts can guide LLMs to approximate gradient-assisted optimization behavior, achieve competitive performance in routing and scheduling problems, and generalize to unseen configurations, all without retraining.
\par
This formulation highlights the fundamental strengths of LLMs as zero-shot or few-shot optimizers capable of learning directly from contextual cues. For UAV networks, this translates to improved operational agility and decision interpretability, since every decision is traceable through natural-language reasoning chains. Such attributes are especially valuable in AVs, where explicit mathematical modeling is complex or infeasible. Integrating LLMs into this domain allows emergency AVs to function as context-aware optimization agents that reason over sensor data, mission goals, and operational constraints simultaneously.
\par
However, several challenges remain before LLM-assisted frameworks can be fully realized in field-deployable UAV systems. These include the quantification of convergence properties, the evaluation of consistency and optimality of inferred decisions, and the integration of multi-modal sensory inputs, such as video, radar, and telemetry, into the prompt context. Moreover, the computational and latency overhead associated with edge or onboard inference must be minimized to meet real-time requirements. Formal convergence analysis of LLM-assisted optimization is particularly challenging due to several intertwined factors. First, LLMs operate in a high-dimensional, non-stationary model space, with billions of parameters and evolving internal states driven by shifting textual contexts. As a result, it is impossible to define a fixed loss function or apply standard Lyapunov-style convergence proofs. Second, iterative updates rely on discrete and stochastic prompt modifications, which are non-differentiable and probabilistic, preventing smooth analytical treatment. Third, environmental uncertainties, such as channel conditions, queue dynamics, and mobility in UAV-assisted data collection, interact with the LLM’s reasoning process, forming a coupled stochastic system that violates the independence and Markov assumptions typically required for convergence analysis. Together, these factors make it analytically intractable to formally establish convergence for LLM-assisted in-context learning or prompt-based optimization systems.
\par
In summary, LLM-assisted ICL provides a lightweight, training-free paradigm for coordinating AVs, complementing the structured optimization of RL and the generative adaptability of DMs. Together, these paradigms form a cohesive foundation for next-generation AVs capable of reasoning, adapting, and optimizing under uncertainty in real-world missions. Overall, DMs enhance DRL by generating synthetic data and improving policy adaptability, yet they remain computationally demanding. In contrast, LLM-assisted optimizers leveraging ICL offer a promising alternative. They enable rapid, training-free adaptation through natural language prompts, iterative refinement, and human-interpretable optimization, key advantages in emergency response situations\cite{emami2025prompts}. Table \ref{tab:comparison} summarizes the comparative performance of these optimization approaches for AVs.

\section{Conclusion} 
\label{sec6}

This paper provides a comprehensive examination of next-generation autonomous emergency response systems, highlighting the complementary roles of DM-augmented RL and LLM-assisted ICL. Together, these paradigms redefine how AVs learn, adapt, and operate amid the uncertainty and time-critical demands of emergency situations. DM-augmented RL enhances policy robustness and data efficiency by generating high-fidelity synthetic experiences that bridge the simulation-to-reality gap, enabling safer and more generalized decision-making in complex environments. However, this performance requires significant computational resources and results in increased inference latency, emphasizing the need for architectural optimization and efficient denoising strategies. In contrast, LLM-assisted ICL offers a lightweight, training-free optimization framework capable of real-time adaptation through contextual prompting and natural language reasoning.This paradigm promotes interpretable, human-aligned decision-making, enabling autonomous systems to dynamically adjust their behaviors without requiring retraining – an invaluable capability in unpredictable emergency operations. Integrating LLMs with DMs offers a promising direction for ITS by combining strong language understanding with high-quality generation. LLMs enable DMs to produce more realistic, context-aware outputs and support intuitive natural language interaction. This synergy can lead to smarter traffic simulation and more accessible, user-driven ITS.

\bibliographystyle{IEEEtran}
\bibliography{references}
\end{document}